\documentclass{llncs}

\usepackage{times}
\usepackage{amsmath}
\usepackage[noend]{algpseudocode}
\usepackage{algorithm}

\newlength{\halftextwidth}
\newcommand{\REGULAR}{\mbox{\sc Regular}}

\newcommand{\myOmit}[1]{}

\newcommand{\CFG} {\mbox{\sc Cfg}}
\newcommand{\WCFG} {\mbox{\sc Wcfg}}
\newcommand{\CYK} {\mbox{\sc CYK}}

\begin{document}

\title{The Weighted $\CFG$ Constraint}

\author{George Katsirelos, Nina Narodytska and Toby Walsh}

\institute{University of New South Wales and
NICTA, Sydney, Australia,
}

\maketitle

\begin{abstract}
We introduce the weighted $\CFG$ constraint and propose a propagation algorithm
that enforces domain consistency in $O(n^3|G|)$ time. We show that this
algorithm can be decomposed into a set of primitive arithmetic constraints without hindering
propagation. 
\end{abstract}
\section {Introduction}

One very promising method for rostering and other domains is to
specify constraints via grammars or automata that accept some
language. We can specify constraints in this 
way on, for instance, the number of consecutive night shifts or
the number of days off in each 7 day period.
With the \REGULAR\ constraint \cite{Pesant04}, we specify
the acceptable assignments to a sequence of variables by 
a deterministic finite automaton. One limitation of this approach
is that the automaton may need to be large. 
For example, there are regular languages which can only be
defined by an automaton with an exponential number
of states. Researchers have therefore
looked higher up the Chomsky hierarchy. In particular, the \CFG\ constraint
\cite{Sellmann06GRAMMAR,qwcp06} permits us to specify constraints using any
context-free grammar. In this paper, we consider
a further generalization to
the weighted \CFG\ constraint. This can model over-constrained
problems and problems with preferences. 

\section{The weighted $\CFG$ constraint}

In a context-free grammar, rules
have a left-hand side with just
one non-terminal, and a right-hand side consisting of
terminals and non-terminals. 
Any context-free grammar can be written
in Chomsky form in which the right-hand
size of a rule is just one terminal or two non-terminals. 
The weighted $\WCFG(G,W,z,[X_1,\ldots,X_n])$ constraint
holds iff an assignment $X$ forms a string 
belonging to the grammar $G$ and 
the minimal weight of a derivation of $X$ 
less than or equal to $z$.  
The matrix $W$ defines weights of productions in 
the grammar $G$.
The weight of a derivation is the sum of production weights used in the derivation.
The \WCFG\ constraint is domain
consistent iff for each variable, every value in its domain
can be extended to an assignment satisfying the constraint.


We give a propagator
for the $\WCFG$ constraint based on an extension
of the $CYK$ parser to probabilistic grammars \cite{Ney91}. 
We assume that $G$ is in Chomsky normal form and
with a single start non-terminal $S$.
The algorithm has two stages. In the first,
we construct a dynamic programing table $V[i,j]$ 
where an element $A$ of $V[i,j]$ is a potential non-terminal
that generates a substring $[X_{i},\ldots,X_{i+j}]$. We compute
a lower bound $l[i,j,A]$ on the minimal weight of a derivation from $A$.
In the second stage, we move from $V[1,n]$ to the bottom of table $V$. 
For an element $A$ of  $V[i,j]$,
we compute an upper bound $u[i,j,A]$ on the maximal weight of a derivation    
from $A$ of a substring $[X_{i},\ldots,X_{i+j}]$. 
We mark the element $A$ iff $l[i,j,A] \leq u[i,j,A]$. 
The pseudo-code is presented in Algorithm~\ref{a:wcyk}.
Lines \ref{a:s_init_ul}--\ref{a:e_init_ul} initialize $l$
and $u$. 
Lines \ref{a:s_up}--\ref{a:e_up} compute the first
stage, whilst lines \ref{a:s_down}--\ref{a:e_down}
compute the second stage. Finally, we prune inconsistent
values 
in lines \ref{a:s_prune}--\ref{a:e_prune}.
Algorithm~\ref{a:wcyk} enforces domain consistency
in $O(|G|n^3)$ time.

\begin{algorithm}
\scriptsize {
\caption{The weighted $\CYK$ propagator}\label{a:wcyk}
\begin{algorithmic}[1]
\Procedure{WCYK-alg}{$G,W,z,[X_1,\ldots,X_n]$}

\For{$j = 1$ \textbf{to} $n$} \label{a:s_init_ul}
	\For{$i =1$ \textbf{to} $n -j + 1$}
		\For{each $A \in G$}
			\State $l[i,j,A] = z+1$; $u[i,j,A] = -1$;  
		\EndFor
	\EndFor
\EndFor \label{a:e_init_ul}

\For{$i = 1$ \textbf{to} $n$ }	 \label{a:s_up}
	\State $V[i,1] = \{A |A \rightarrow a \in G, a \in D(X_i)\}$
	\For{$ A \in V[i,1]$ s.t $ A \rightarrow a \in G, a \in D(X_i)$} \label{a:first_row_cond} 	
		\State $l[i,1,A]  = \min\{l[i,1,A], W[A \rightarrow a]\}$;
	\EndFor			
\EndFor

\For{$j = 2$ \textbf{to} $n$ } \label{a:j_down}
	\For{$i =1$ \textbf{to} $n -j + 1$}
		\State $V[i,j] = \emptyset $;
		\For{$k =1$ \textbf{to} $j - 1$} \label{a:k_down}
				\State $V[i,j] = V[i,j] \cup \{A|A \rightarrow BC \in G, B \in V[i,k], C \in V[i+k,j-k]\}$
				\For{each $A \rightarrow BC \in G$ s.t. $B \in V[i,k], C \in V[i+k,j-k]$}
						\State $l[i,j,A] = \min \{l[i,j,A], W[A \rightarrow BC] + l[i,k,B] + l[i+k,j-k,C]\}$; \label{a:cal_min}
				\EndFor			
		\EndFor
	\EndFor
\EndFor\label{a:e_up}

\If {$S \ \notin V[1,n]$} \State return 0;
\EndIf	

\State mark $(1,n,S)$; $u[1,n,S] = z$;
\For{$j =n$ \textbf{downto} $2$ } \label{a:s_down} \label{a:j_up}
	\For{$i =1$ \textbf{to} $n -j + 1$}
		\For{$A$ such that $(i,j,A)$ is marked}
					\For{$k = 1$ \textbf{to} $j - 1$}	\label{a:k_up}					
						\For{each $A \rightarrow BC \in G$ s.t. $B \in V[i,k], C \in V[i+k,j-k]$}				
								\If {$W[A \rightarrow BC] + l[i,k,B] +  l[i+k,j-k,C] > u[i,j,A]$} \label{a:cond2}
										\State continue;			
								\EndIf	
								\State mark $(i, k, B)$; mark $(i+k, j-k, C)$;								
								\State $u[i,k,B] = \max \{u[i,k,B], u[i,j,A] - l[i+k,j-k,C] - W[A \rightarrow BC]\}$; \label{a:cal_max1}
								\State $u[i+k,j-k,C] = \max \{u[i+k,j-k,C], u[i,j,A] - l[i,k,B] - W[A \rightarrow BC]\}$; \label{a:cal_max2}
				\EndFor				
			\EndFor			
		\EndFor
	\EndFor
\EndFor \label{a:e_down}

\For{$i =1$ \textbf{to} $n$}\label{a:s_prune}
	\State $D(X_i) = \{a \in D(X_i)| A \rightarrow a \in G, (i,1,A)\ is\ marked\ and\ W[A \rightarrow a] \leq u[i,1,A]\}$;
\EndFor\label{a:e_prune}

\State return 1;
\EndProcedure
\end{algorithmic}
}
\end{algorithm}

\section {Decomposition of the weighted $\CFG$ constraint}
\label{s:wcfg_dec}
As an alternative to this monolithic propagator,
we propose a simple decomposition with which
we can also enforce domain consistency.
A decomposition has several advantages.  For example, 
it is easy to add to any constraint solver. 
As a second example, decomposition gives an
efficient incremental propagator, and opens the
door to advanced techniques like nogood learning
and watched literals. The idea of the decomposition is to introduce
arithmetic constraints to compute $l$ and $u$.
Given the 
table $V$ obtained by Algorithm~\ref{a:wcyk},
we construct the corresponding $AND/OR$ directed acyclic graph (DAG) 
as in~\cite{Quimper07GRAMMAR}. 
We label an $OR$ node  by $n(i,j,A)$,
and an $AND$ node by $n(i,j,k,A \rightarrow BC)$.
We denote the parents of a node $nd$ as $PRT(nd)$
and the children as $CHD(nd)$. For each node two
integer variables are introduced to compute
$l$ and $u$. 
For an $OR$-node $nd$, these are
$l_{O}(nd)$ and $u_{O}(nd)$, whilst for an $AND$-node  
$nd$, these are $l_{A}(nd)$, $u_{A}(nd)$.
 
For each $AND$ node $nd= n(i,j,k,A \rightarrow BC)$ we post a constraint
to connect $nd$ to its children $CHD(nd)$:
\begin{gather}
l_{A}(nd) = \sum_{n_{c} \in CHD(nd)} {{l_{O}(n_{c})}} +  W[A \rightarrow BC] \label{e:and_sum}
\end{gather}
For each $OR$ node $nd= n(i,j, A)$ we post constraints
to connect $nd$ to its children $CHD(nd)$:

\begin{gather}
l_{O}(nd) = \min_{n_{c} \in CHD(nd)}  \{l_{A}(n_c)\} \label{e:or_min} \\
u_{O}(nd) = u_{A}(n_c),\ n_c \in CHD(nd) \label{e:or_max_1} 
\end{gather}

For each $OR$ node $nd= n(i,j, A)$ we post a set of constraints
to connect $nd$ to its parents $PRT(nd)$ and siblings:
\begin{gather}
u_{O}(nd) = max_{n_{p} \in PRT(nd)}  \{u_{A}(n_p) - l_{O}(n_{sb}) - W[P]\} \label{e:or_prt_max},  
\end{gather}
where $P =B \rightarrow AC$ or $B \rightarrow CA$,  $n_p = n(r,q,t,P)$ is the parent of $nd = n(i,j, A)$ and $n_{sb} = 	n(i_1,j_1,C)$.

Finally, we introduce constraints to prune $X_i$. 
For each leaf of the DAG that is an $OR$ node $nd = n(i,1,a)$, we introduce:
\begin{gather}
 a \in D(X_i)  \Rightarrow  0 \leq l_{O}(nd) \leq z \label{e:prune_1} \\
 a \notin D(X_i)  \Leftrightarrow  l_{O}(nd) > z \label{e:prune_3} \\
 l_{O}(nd) > u_{O}(nd) \Rightarrow a \notin D(X_i) \label{e:prune_5}
\end{gather} 

As the maximal weight of a derivation is less than or equal to $z$ we post:
\begin{gather}
u_{O}(n(1,n,S)) \leq z \label{e:prune_8} 
\end{gather}

Bounds propagation will set
the lower bound of $l_{O}(n(i,j,A))$ to the minimal weight of 
a derivation from $A$, and 
the upper bound on $u_{O}(n(i,j,A))$
to the maximum weight of a derivation
from $A$. 
We forbid  branching on variables $l_{A|O}$ and
$u_{A|O}$ as branching on $l_{A|O}$
would change the weights matrix $W$ and branching on
$u_{A|O}$ would add additional
restrictions to the weight of a derivation. 
Bounds propagation 
on this decomposition enforces domain consistency
on the \WCFG\ constraint. 
If we invoke constraints in the decomposition
in the same order as we compute the table $V$,
this takes $O(n^3|G|)$ time.
For simpler grammars, propagation is faster.
For instance, as in the unweighted
case, it takes just $O(n|G|)$ time
on a regular grammar. 

We can speed up propagation by recognizing when 
constraints are entailed. 
If  $l_{O}(nd)$ $>$ $ u_{O}(nd)$ holds for an $OR$ node $nd$ then
constraints~\eqref{e:or_prt_max} and~\eqref{e:or_min}
are entailed. If  $l_{A}(nd) > u_{A}(nd)$ holds for an $AND$ node $nd$ then
constraints~\eqref{e:and_sum} and~\eqref{e:or_max_1} 
are entailed. 
To model entailment we augmented each of these constraints 
in such a way that if $l_{O}(nd) > u_{O}(nd)$ or
$l_{A}(nd)$ $>$ $u_{A}(nd)$ hold then corresponding
constraints are not invoked by the solver.

\section {The Soft $\CFG$ constraint}
\label{s:wcfg_soft}
We can use the \WCFG\ constraint to encode
a soft version of $\CFG$ constraint
which is useful for modelling over-constrained problems. 
The soft  $\CFG(G,z,[X_1,\ldots,X_n])$ constraint holds iff
the string $[X_1,\ldots,X_n]$ is at most distance $z$ from
a string in $G$. We consider both
Hamming and edit distances. 
We encode the soft $\CFG(G,z,[X_1,\ldots,X_n])$ 
constraint as a weighted $\CFG(G',W,z,[X_1,\ldots,X_n])$ 
constraint.  For Hamming
distance, for each production $A \rightarrow a \in G$, 
we introduce additional unit weight
productions to simulate substitution:
\begin{gather*}
\{A \rightarrow  b, W[A \rightarrow  b] = 1| A \rightarrow a \in G, A \rightarrow b \notin G,  b \in \Sigma  \}
\end{gather*} 
Existing productions have zero weight.
For edit distance, we introduce 
additional productions to simulate substitution, 
insertion and deletion:
\begin{gather*}
 \{A \rightarrow  b, W[A \rightarrow  b] = 1| A \rightarrow a \in G, A \rightarrow b \notin G,  b \in \Sigma \}\cup \\
		\{A \rightarrow \varepsilon, W[A \rightarrow  \varepsilon] = 1|  a \in \Sigma\}\cup\\
    \{A \rightarrow  Aa, W[A \rightarrow  Aa] = 1|  a \in \Sigma\}\cup  \\
    \{A \rightarrow  aA, W[A \rightarrow  aA] = 1|  a \in \Sigma\}
\end{gather*} 
To handle $\varepsilon$  productions we 
modify Alg.~\ref{a:wcyk} so loops in lines~\eqref{a:k_down},\eqref{a:k_up} run from $0$ to $j$.

\section{Experimental results}
We evaluated these propagation methods on 
shift-scheduling benchmarks~\cite{Demassey05,Claude07}. 
A personal schedule is
subject to various regulation rules, e.g. a full-time employee
has to have a one-hour lunch. 
This rules are encoded into a context-free grammar augmented 
with restrictions on productions~\cite{Quimper07GRAMMAR,Quimper2007LNS}.
A schedule for an employee has $n = 96$ slots represented
by $n$ variables. In each slot,
an employee can work on an activity ($a_i$), take a break ($b$), lunch ($l$) or  
rest ($r$). These rules are represented by the following 
grammar:
\begin{displaymath}
\begin{array}{ccc}
	S\rightarrow RPR,   f_P(i,j)\equiv 13 \leq j \leq 24,  \ &  P \rightarrow WbW, & L \rightarrow lL|l,   f_L(i,j)\equiv j = 4   \\
	S\rightarrow RFR,   f_F(i,j)\equiv 30 \leq j \leq 38,  \ & R \rightarrow rR|r, 	& W \rightarrow A_i,  f_W(i,j)\equiv j \geq 4 \\
	A_i \rightarrow a_iA_i|a_i, f_A(i,j)\equiv open(i), \ & F \rightarrow PLP  & \\
\end{array}
\end{displaymath}
where functions $f(i,j)$ are restrictions on productions and $open(i)$ is a function that returns 
$1$ if the business is opened at $i$th slot and $0$ otherwise. 
To model  labour demand for a slot we introduce Boolean variables $b(i,j,a_k)$,
equal to $1$ if $j$th employee performs 
activity $a_k$ at $i$th time slot. For each time slot $i$ and activity $a_k$ we post a constraint 
$\sum_{j=1}^m x(i,j,a_k) > d(i,a_k)$, where $m$ is the number of employees. 
The goal is
to minimize the number of slots in which employees worked.

We used Gecode 2.0.1 for our experiments and
ran them on an Intel Xeon $2.0$Ghz with $4$Gb of RAM
\footnote{We would like to thank Claude-Guy Quimper for his help with the experiments}. 
In the first set of experiments, we used the
weighted $\CFG(G,z_j,X)$, $j=1,\ldots,m$ 
with zero weights.
Our monolithic propagator gave similar results to 
the unweighted \CFG\ propagator from \cite{Quimper07GRAMMAR}.
Decompositions 
were slower than decompositions of the unweighted \CFG\ 
constraint as the former uses integers
instead of Booleans. 
In the second set of experiments,
we assigned weight $1$ to activity productions, like 
$A_i \rightarrow a_i$, and post an additional cost function $\sum_{j=1}^m z_j$ that is minimized. $\sum_{j=1}^m z_j$ is 
the  number of slots  in which employees worked. 
Results are presented in Table\ref{t:t1}.
We improved on the best solution found in the first model 
in $4$ benchmarks and proved optimality in one. 
The decomposition of the weighted $\CFG$ constraint was slightly 
slower than the monolithic propagator,
while entailment improved performance in most cases.

\vspace*{-12pt}
\begin{table}
\begin{center}
{\scriptsize
\begin{tabular}{|ccc|cccc|cccc|cccc|c|c|}
\hline
&
&
&\multicolumn {4}{|c|}{Monolithic}
&\multicolumn {4}{|c|}{Decomposition}
&\multicolumn {4}{|c|}{Decomption+entailment}
&\multicolumn {1}{|c|}{}
&\multicolumn {1}{|c|}{}\\
 \hline
$|A|$& $\#$& m & cost & time & bt & BT& cost & time & bt & BT& cost & time & bt  &BT & Opt & Imp\\ 
\hline
 $1$ & $2$ & $4$ & \textbf{107} & \textbf{5} & \textbf{0} &  8652 & \textbf{107} &  7 & \textbf{0} &  5926 & \textbf{107} &  7 & \textbf{0} & \textbf{11521} & &\\
 $1$ & $3$ & $6$ & \textbf{148} & \textbf{7} & \textbf{1} &  5917 & \textbf{148} &  34 & \textbf{1} &  1311 & \textbf{148} &  9 & \textbf{1} & \textbf{8075} & &\\
 $1$ & $4$ & $6$ & \textbf{152} &  1836 & \textbf{5831} &  11345 & \textbf{152} & \textbf{1379} & \textbf{5831} & \textbf{14815} & \textbf{152} &  1590 & \textbf{5831} &  13287 & &\\
 $1$ & $5$ & $5$ & \textbf{96} &  6 & \textbf{0} &  8753 & \textbf{96} &  6 & \textbf{0} &  2660 & \textbf{96} & \textbf{3} & \textbf{0} & \textbf{45097} & &\\
 $1$ & $6$ & $6$ &  $ - $ &  $ - $ &  $ - $ &  10868 & \textbf{132} &  3029 & \textbf{11181} &  13085 & \textbf{132} & \textbf{2367} & \textbf{11181} & \textbf{16972} & &\\
 $1$ & $7$ & $8$ & \textbf{196} &  16 & \textbf{16} &  10811 & \textbf{196} &  18 & \textbf{16} &  6270 & \textbf{196} & \textbf{15} & \textbf{16} & \textbf{10909} & &\\
 $1$ & $8$ & $3$ & \textbf{82} &  11 & \textbf{9} & \textbf{66} & \textbf{82} &  13 & \textbf{9} & \textbf{66} & \textbf{82} & \textbf{5} & \textbf{9} & \textbf{66} & $\surd$&$\surd$\\
 $1$ & $10$ & $9$ &  $ - $ &  $ - $ &  $ - $ &  10871 &  $ - $ &  $ - $ &  $ - $ &  9627 &  $ - $ &  $ - $ &  $ - $ & \textbf{18326} & &\\
\hline
 $2$ & $1$ & $5$ & \textbf{100} &  523 & \textbf{1109} &  7678 & \textbf{100} &  634 & \textbf{1109} &  6646 & \textbf{100} & \textbf{90} & \textbf{1109} & \textbf{46137} & &\\
 $2$ & $2$ & $10$ &  $ - $ &  $ - $ &  $ - $ & \textbf{11768} &  $ - $ &  $ - $ &  $ - $ &  10725 &  $ - $ &  $ - $ &  $ - $ &  6885 & &\\
 $2$ & $3$ & $6$ & \textbf{165} &  3517 &  \textbf{9042} &  9254 &  168 & 2702 & 4521 &  6124 & \textbf{165} &  \textbf{2856} &  \textbf{9042} & \textbf{11450} & &$\surd$\\
 $2$ & $4$ & $11$ &  $ - $ &  $ - $ &  $ - $ & \textbf{8027} &  $ - $ &  $ - $ &  $ - $ &  6201 &  $ - $ &  $ - $ &  $ - $ &  5579 & &\\
 $2$ & $5$ & $4$ & \textbf{92} & \textbf{37} & \textbf{118} & \textbf{12499} & \textbf{92} &  59 & \textbf{118} &  6332 & \textbf{92} &  49 & \textbf{118} &  10329 & &\\
 $2$ & $6$ & $5$ & \textbf{107} & \textbf{9} & \textbf{2} &  6288 & \textbf{107} &  22 & \textbf{2} &  1377 & \textbf{107} &  14 & \textbf{2} & \textbf{7434} & &\\
 $2$ & $8$ & $5$ & \textbf{126} &  422 & \textbf{1282} &  12669 & \textbf{126} &  1183 & \textbf{1282} &  3916 & \textbf{126} & \textbf{314} & \textbf{1282} & \textbf{16556} & &$\surd$\\
 $2$ & $9$ & $3$ & \textbf{76} &  1458 & \textbf{3588} &  8885 & \textbf{76} &  2455 & \textbf{3588} &  5313 & \textbf{76} & \textbf{263} & \textbf{3588} & \textbf{53345} & &$\surd$\\
 $2$ & $10$ & $8$ &  $ - $ &  $ - $ &  $ - $ &  3223 &  $ - $ &  $ - $ &  $ - $ &  3760 &  $ - $ &  $ - $ &  $ - $ & \textbf{8827} & &\\
\hline
\end{tabular}}
\caption{\label{t:t1} 
All benchmarks have one-hour time limit.  $|A|$ is the number of activities,$m$ is the number of employees, $cost$ shows the total number of slots in which employees worked in the best solution,  $time$ is the time to find the best solution, $bt$ is the number of backtracks to find the best solution,  $BT$ is the number of backtracks in one hour, $Opt$ shows if optimality is proved, $Imp$ shows if a lower cost solution is found by the second model}
\end{center}
\end{table}
\vspace*{-18pt}
\bibliographystyle{splncs}

\end{document}